\documentclass[10pt,twocolumn,letterpaper]{article}

\usepackage{cvpr}
\usepackage{times}
\usepackage{epsfig}
\usepackage{graphicx}
\usepackage{amsmath}
\usepackage{amssymb}

\usepackage{booktabs}
\usepackage{bm}
\usepackage{enumerate}
\usepackage{amsfonts}
\usepackage{amsmath}
\usepackage{tabularx}
\usepackage{caption}
\usepackage{graphicx}
\usepackage{subcaption}
\usepackage{amssymb}
\usepackage[para,online,flushleft]{threeparttable}
\usepackage{color}
\usepackage{subfiles}

\newcommand{\mymathfont}{}

\usepackage[pagebackref=true,breaklinks=true,letterpaper=true,colorlinks,bookmarks=false]{hyperref}

\cvprfinalcopy %

\def\cvprPaperID{6694} %

\ifcvprfinal\fi
\begin{document}

\title{Progressive Relation Learning for Group Activity Recognition}

\author{Guyue Hu\textsuperscript{\rm 1,3}\thanks{Corresponding Author: Guyue Hu} ,
	Bo Cui\textsuperscript{\rm 1,3},
	Yuan He\textsuperscript{\rm 1,3},
	Shan Yu\textsuperscript{\rm 1,2,3}\\
	\textsuperscript{\rm 1}Brainnetome Center, National Laboratory of Pattern Recognition (NLPR), \\
	Institute of Automation, Chinese Academy of Sciences (CASIA)\\
	\textsuperscript{\rm 2}Center for Excellence in Brain Science and Intelligence Technology (CEBSIT)\\
	\textsuperscript{\rm 3}University of Chinese Academy of Sciences (UCAS)\\
	{\tt\small \{guyue.hu, bo.cui, yuan.he, shan.yu\}@nlpr.ia.ac.cn}
}

\maketitle

\begin{abstract}
	Group activities usually involve spatiotemporal dynamics among many interactive individuals, while only a few participants at several key frames essentially define the activity. Therefore, effectively modeling the group-relevant and suppressing the irrelevant actions (and interactions) are vital for group activity recognition. In this paper, we propose a novel method based on deep reinforcement learning to progressively refine the low-level features and high-level relations of group activities. Firstly, we construct a semantic relation graph (SRG) to explicitly model the relations among persons. Then, two agents adopting policy according to two Markov decision processes are applied to progressively refine the SRG. Specifically, one feature-distilling (FD) agent in the discrete action space refines the low-level spatiotemporal features by distilling the most informative frames. Another relation-gating (RG) agent in continuous action space adjusts the high-level semantic graph to pay more attention to group-relevant relations. The SRG, FD agent, and RG agent are optimized alternately to mutually boost the performance of each other. Extensive experiments on two widely used benchmarks demonstrate the effectiveness and superiority of the proposed approach.
\end{abstract}

\section{Introduction}
Group activity %
recognition, which refers to discern the activities involving a large number of interactive individuals, %
has attracted growing interests in the communities of computer vision \cite{DBLP:conf/cvpr/DengVHM16,DBLP:conf/cvpr/WangNY17,tang2018mining,yan2018participation,DBLP:conf/eccv/QiQLWLG18}. %
Unlike conventional video action recognition that only concentrates on the spatiotemporal dynamics of one or two persons, group activity recognition further requires understanding the group-relevant interactions among many individuals. 

\begin{figure}[tbp]
	\centering
	\includegraphics[width=0.95\linewidth,height=0.89\linewidth]{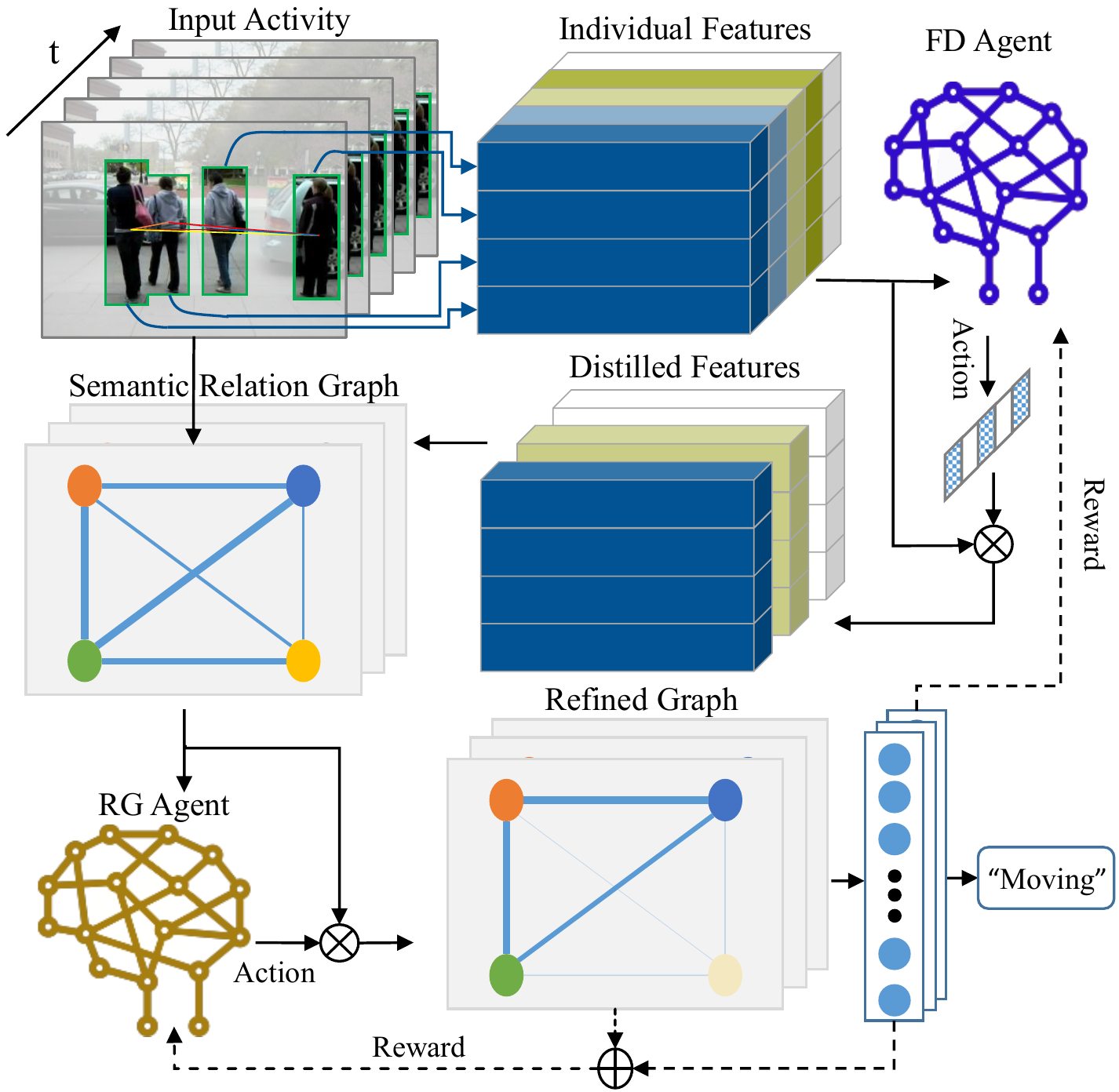}
	\caption{The overview of proposed method. A feature-distilling (FD) agent progressively selects the most informative frames of the low-level spatiotemporal individual features. A relation-gating (RG) agent further progressively refines the high-level semantic relation graph (SRG) to discover group-relevant relations.}
	\label{overview}
\end{figure}

In the past a few years, a series of approaches combine the hand-crafted feature with probability graph \cite{choi2012unified,DBLP:journals/pami/LanWYRM12,DBLP:conf/cvpr/ShuXRTZ15}. %
Recently, the LSTM, strucural RNNs and message passing neural network (MPNN) are also applied to model the interactions among persons, subgroups and groups \cite{DBLP:conf/eccv/QiQLWLG18,DBLP:conf/cvpr/WangNY17,DBLP:conf/wacv/BiswasG18}. The interaction relations in these methods are \textit{implicitly} contained in the ordered RNNs or the passing messages of MPNN. Moreover, not all the existing relations are relevant to the group activity and the pairwise relations may contain many edges that are coupled from spurious noise, such as cluttered background, inaccurate human detection, and interaction between outlier persons (\eg, the ``Waiting" person in Fig.~\ref{overview}). Due to the relations in previous methods are modeled \textit{implicitly}, it is unable to determine whether one specific relation is group-relevant or not.

In addition, although a large number of persons may involve in a group activity, usually only a few actions or interactions in several key frames essentially define the group activity. Yan \etal~\cite{yan2018participation} heuristically defined the key participants as the ones with ``long motion" and ``flash motion". Qi \etal~\cite{DBLP:conf/eccv/QiQLWLG18} applied a ``self-attention" mechanism to attend to important persons and key frames. %
Nevertheless, these methods are limited to the coarse individual (person) level, and have not dug into the fine-grained relation level to consider which relations are vital (\eg, regulating 15 pairwise relations is more fine-grained than attending 6 persons). %

To move beyond such limitations, we propose a progressive relation learning framework to effectively model and distill the group-relevant actions and interactions in group activities. Firstly, we build a graph to explicitly model the semantic relations in group activities. %
Then, as illustrated in Fig.~\ref{overview}, two agents progressively refine the low-level spatiotemporal features and high-level semantic relations of group activities. Specifically, at the feature level, a feature-distilling agent explores a policy to distill the most informative frames of low-level spatiotemporal features. At the relation level, a relation-gating agent further refines the high-level %
relation graph to focus on the group-relevant relations. %

In summary, the contributions of this paper can be summarized as: (1) A novel progressive relation learning framework is proposed for group activity analysis. (2) Beyond distilling group-relevant information at the course individual (person) level, we proposed a RG agent to progressively discover group-relevant semantic relations at the fine-grained relation level. (3) A FD agent is proposed to further progressively filter the frames of low-level spatiotemporal features that used for constructing the high-level semantic relation graph.

\section{Related Works}
\textbf{Reinforcement Learning}. Reinforcement learning (RL) has benefited many fields of computer vision, such as image cropping \cite{DBLP:conf/cvpr/LiWZH18} %
and visual semantic navigation \cite{DBLP:journals/corr/abs-1810-06543}. %
Regarding the optimization policy, RL can be categorized into the value-based methods, policy-based methods, and their hybrids. The value-based methods (\eg, deep Q-learning \cite{DBLP:journals/corr/MnihKSGAWR13}) are good at solving the problems in low dimensional discrete action space, but they fail in high dimensional continuous space. Although the policy-based methods (\eg, policy gradient \cite{DBLP:conf/nips/SuttonMSM99}) %
are capable to deal with the problems in continuous space, they suffer from high variance of gradient estimation. The hybrid methods, such as Actor-Critic algorithms \cite{DBLP:conf/nips/KondaT99}, combine their advantages and are capable for both of discrete and continuous action spaces. Moreover, by exploiting %
asynchronous updating, the Asynchronous Advantage Actor-Critic (A3C) algorithm \cite{DBLP:conf/icml/MnihBMGLHSK16} has largely improved the training efficiency. Therefore, we adopt the A3C algorithm to optimize both of our RG agent in continuous action space and our FD agent in discrete action space.

\textbf{Graph Neural Network}. Due to the advantages of representing and 
reasoning over structured data, the graph neural network (GNN) has attracted increasing attention \cite{DBLP:journals/corr/abs-1810-00826,DBLP:journals/corr/abs-1901-00596,DBLP:journals/icme/gyhu,hu2019joint,DBLP:journals/corr/abs-1806-01261}. %
Graph convolutional network (GCN) generalizes CNN on graph, which therefore can deal with non-Euclidean data \cite{DBLP:journals/spm/BronsteinBLSV17}. It has been widely applied in computer vision, \eg, point cloud classification \cite{DBLP:conf/cvpr/SimonovskyK17}, action recognition \cite{DBLP:conf/aaai/YanXL18}, and traffic forecasting \cite{DBLP:conf/ijcai/YuYZ18}. 
Another class of GNN combines graph with RNN, in which each node captures the semantic relation and structured information from its neighbors through multiple iterations of passing and updating, \eg, message-passing neural network \cite{DBLP:conf/icml/GilmerSRVD17}, graph network block \cite{DBLP:conf/icml/Sanchez-Gonzalez18}. %
Each relation in the former class (\ie, GCN) is represented by a scalar in its adjacency matrix that is not adequate for modeling the complex context information in group activity. Therefore, our semantic relation graph is built under the umbrella of the latter class that each relation is explicitly represented by a learnable vector.

\begin{figure*}[tbp]
	\centering
	\includegraphics[width=0.95\linewidth,height=0.45\linewidth]{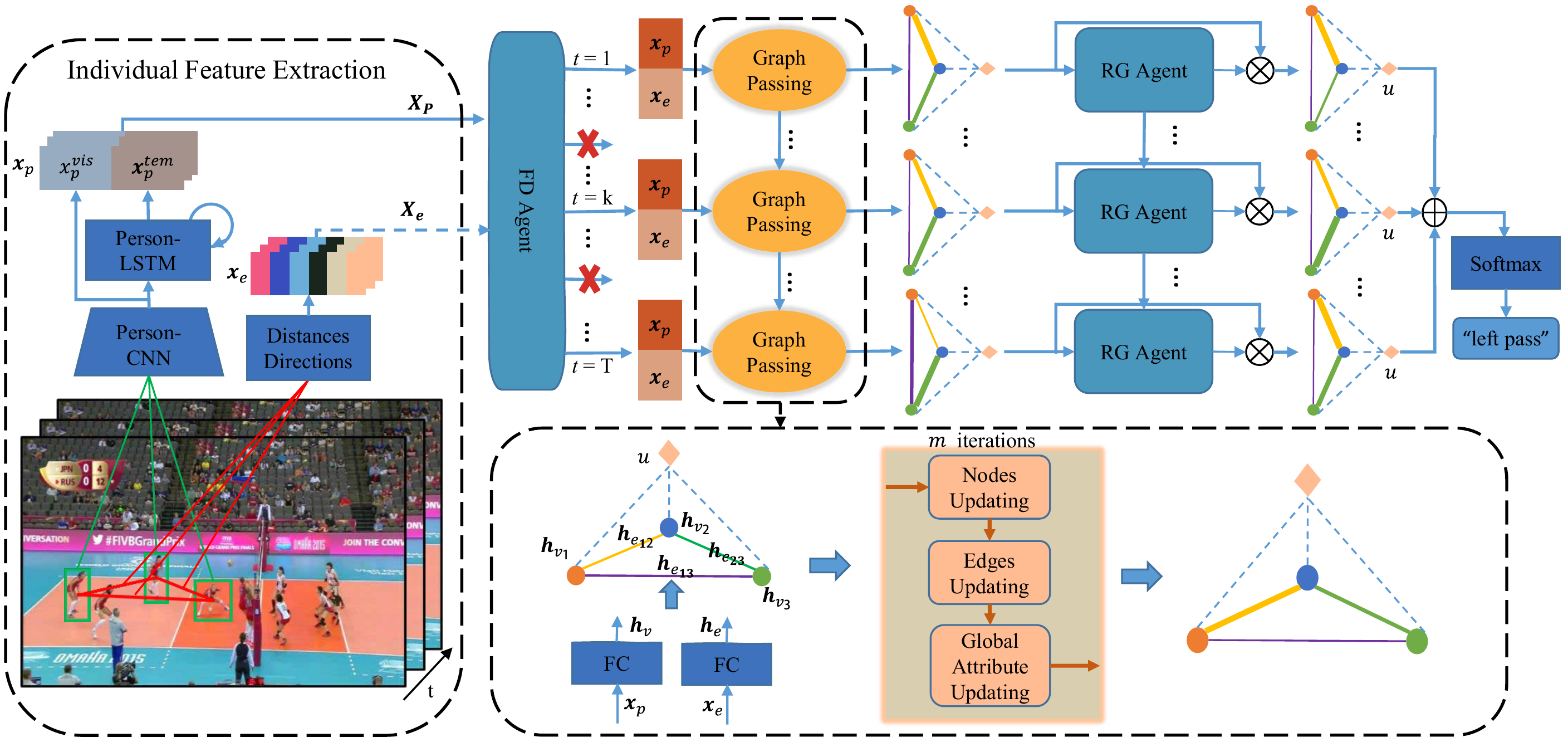}
	\caption{The detailed framework of our method. The low-level spatiotemporal
		features of persons are extracted by a CNN and a LSTM. The feature-distilling (FD) agent selects the informative %
		frames of features. Then the distilled features are used to build a high-level semantic relation graph (SRG), and a relation-gating (RG) agent further refines the SRG. ``FC" denotes fully connected layer. Finally, the activity category is predicted according to the sum of global attributes at all the times.}
	\label{framework}
\end{figure*}

\section{Method}

\subsection{Individual Feature Extraction}
Following \cite{yan2018participation}, the person bounding boxes are firstly obtained through the object tracker %
in the Dlib library \cite{DBLP:journals/jmlr/King09}.
As shown in Fig.~\ref{framework}, %
the visual feature (\eg, appearance and pose) $ x^{vis}_{p_i}$ of each person $ i $ is extracted through a convolutional neural network (called Person-CNN). Then, the spatial visual feature is fed into a long short-term memory network (called Person-LSTM) to model the individual temporal dynamic $ x^{tem }_{p_i} $. Finally, we concatenate the stacked visual features $ \bm{x}^{vis}_{p} $ and temporal dynamics $ \bm{x}^{tem}_{p} $ of all persons as the basic spatiotemporal features, \ie, $ \bm{x}_p =[\bm{x}^{vis}_{p},\bm{x}^{tem}_{p}]$. %
These basic representations %
contain no context information, such as the person to person, person to group, and group to group interactions. %
Besides, the spatial distance vectors $ \{ |dx|, |dy|, |dx+dy|, \sqrt{(dx)^2+(dy)^2}\}$  and direction vectors $\{arctan(dy,dx), arctan2(dy,dx)\} $ between each pair of persons are concatenated as the original interaction features $ \bm{x_e} $, where $ dx $ and $ dy $ are the displacements along horizontal and vertical axes, respectively.

\subsection{Semantic Relation Graph}
Inferring semantic relations over inherent structure in a scene is helpful to suppress noises, such as inaccurate human detection, mistaken action recognition, and outlier people not involved in a particular group activity. To achieve it, we explicitly model the structured relations through a graph network \cite{DBLP:conf/icml/Sanchez-Gonzalez18}. Let us put aside the two agents in Fig.~\ref{framework} and explain how to build the baseline semantic relation graph first. Let a graph $\bm{G} = (\bm{u}, \bm{V}, \bm{E})$ , where $ \bm{u} $ is the global attribute (\ie, activity score),  $\bm{V}=\{v_i\}_{i=1}^{N_v} $ and $ \bm{E}=\{e_{ij}\}_{i,j=1}^{N_v}$ are respectively the person nodes and the relation edges among them. The attributes of person nodes $ \bm{H_{v}} $ and the attributes of relation edges $ \bm{H_{e}} $ are respectively initialized with the embeddings of low-level spatiotemporal features  $ \bm{X_p} $ and original interaction features $ \bm{X_e} $.

During graph passing,  each node $ v_i $ collects the contextual information $ \bm{h_{ve}^{ij}} $ from each of its neighbors $ v_j $ ($j \in \mathcal{N}(v_i) $) via a collecting function $ \phi_{ve} $, and aggregates all collected information via an aggregating function $ \psi_{v} $, \ie, 
\begin{equation}
	\bm{h_{ve}}^{ij}
	= \phi_{ve}(\bm{h_{e_{ij}}},\bm{h_{v_j}}) 
	= \text{NN}_{ve}\left([\bm{h_{e_{ij}}},\bm{h_{v_j}}]\right) 
\end{equation}
\begin{equation}
	\bm{\overline{h}_{e_{i}}}
	= \psi_{v}  (\bm{h_{ve}}^{i})=\sum_{j\in\mathcal{N}(v_i)}\bm{h_{ve}}^{ij}
\end{equation}
where the collecting function $ \phi_{ve} $ is %
implemented by a neural network $ \text{NN}_{ve} $, and [$ \cdot $] denotes concatenation.
Then, the aggregated contextual information $ \bm{\overline{h}_{e_{i}}} $ updates the node attributes via a node updating function $ \phi_{v} $ (network $ \text{NN}_v $), 
\begin{equation}
	\bm{h'_{v_i}}
	= \phi_{v}(\bm{\overline{h}_{e_{i}}},\bm{h_{v_{i}}}) 
	= \text{NN}_{v}\left([\bm{\overline{h}_{e_{i}}},\bm{h_{v_{i}}}]\right). 
\end{equation}
After that, each edge $ \bm{h_{e_{ij}}} $ enrolls  message from the sender $ \bm{h'_{v_i}} $ and receiver $ \bm{h'_{v_j}} $ to update its edge attributes via an edge updating function $ \phi_e $ (network $ \text{NN}_e $), 

\begin{equation}
	\bm{\hat{h}}_{\bm{e}_{ij}}
	= \phi_e(\bm{h_{v'_i}}, \bm{h_{v'_j}}, \bm{h_{e_{ij}}})
	= \text{NN}_e\left([\bm{h_{v'_i}}, \bm{h_{v'_j}}, \bm{h_{e_{ij}}}]\right)
\end{equation}
To simplify the problem, we consider the graph is undirected (\ie,
$ \bm{{h'}_{e_{ij}}}=\bm{{h'}_{e_{ji}}} = (\bm{\hat{h}_{e_{ij}}}+\bm{\hat{h}_{e_{ji}}})/2 $) and has no self-connection. %
Finally, the global attribute $ \bm{u} $ is updated based on semantic relations in the whole %
relation graph, \ie,
\begin{equation}
	\bm{u'}
	= \bm{W_u} \left( \sum_{i=1}^{N_v}\sum_{j > i}^{N_v} \bm{h'_{e_{ij}}}\right)  + \bm{b_u}
\end{equation}
where $ \bm{W_u} $ is parameter matrix and $\bm{b_u}$ is bias. The $ \text{NN}_e $, $ \text{NN}_{ve} $ and $ \text{NN}_v $ are implemented with LSTM networks.

Since propagating information over the graph once captures at most pairwise relations, we %
update the graph for $ \bm{m} $ iterations to encode high-order interactions. %
After the propagations, the graph automatically learns the high-level semantic relations %
from the low-level individual features in the scene. Finally, the activity score can be obtained by appending a softmax layer to the $ \bm{u} $ after the last iteration.

\subsection{Progressively Relation Gating}
Although the above fully-connected semantic graph is capable of explicitly modeling any type of relation, it contains many group-irrelevant relations. Therefore, we introduce a relation-gating agent to explore an adaptive policy to select group-relevant relations. %
The decision process is formulated as a Markov Process $ \mathcal{M}=\{S, A, \mathcal{T}, r, \gamma \}$. %

\textbf{States.} The state $ S $ consists of three parts $S=\{S_g,S_l,S_u\}$. $ S_g $ is the whole semantic graph, represented by the stack of all relation triplets (\textit{``sender", ``relation", ``receiver"}), which provides the global information about the current scene. $ S_l $ is concatenation of the relation triplet $(\bm{h}_{\bm{v}_i},\bm{h}_{\bm{e}_{ij}},\bm{h}_{\bm{v}_j}) $ corresponding to one specific relation $ \bm{h}_{\bm{e}_{ij}} $ that will be refined, which provides the local information for the agent. 
$ S_l\in \mathbb{R}^{D_v+D_e+D_v} $, where $ D_v$ and $ D_e $ denote the attribute dimensions of $ N_v $ nodes and $ N_e $ relations, respectively.  $ S_u = \bm{u} $ is global attributes of the relation graph at the current state, where $ \bm{u} $ is the activity scores.

\textbf{Action.} Inspired by the information gates in the LSTMs, we introduce a gate  $ g_{ij} $ for each relation edge. The action $ A $ of the agent is to generate %
the gate $ g_{ij} \in [0,1] $. Then, it is applied to adjust the corresponding relation at each reinforcement step, \ie, $ \bm{h}_{\bm{e}_{ij}}$ = $g_{ij} \cdot \bm{h}_{\bm{e}_{ij}} $. 
Since the semantic relation graph is undirected, we normalize the values of gates before gating operation, \ie, $ g_{ij}=g_{ji}=(g_{ij}+g_{ji})/2 $.    

\begin{figure}[tbp]
	\begin{subfigure}[b]{0.3\linewidth}
		\includegraphics[width=1\linewidth,height=2.9in]{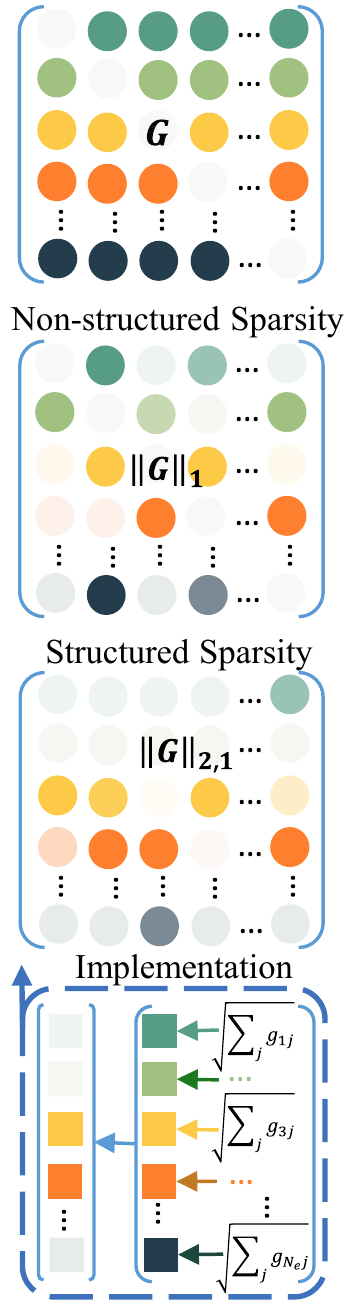}
		\caption{Comparison of sparsity}
		\label{RG:a}
	\end{subfigure}
	\hspace{0.05\linewidth}
	\begin{subfigure}[b]{0.6\linewidth}
		\includegraphics[width=0.95\linewidth,height=3.0in]{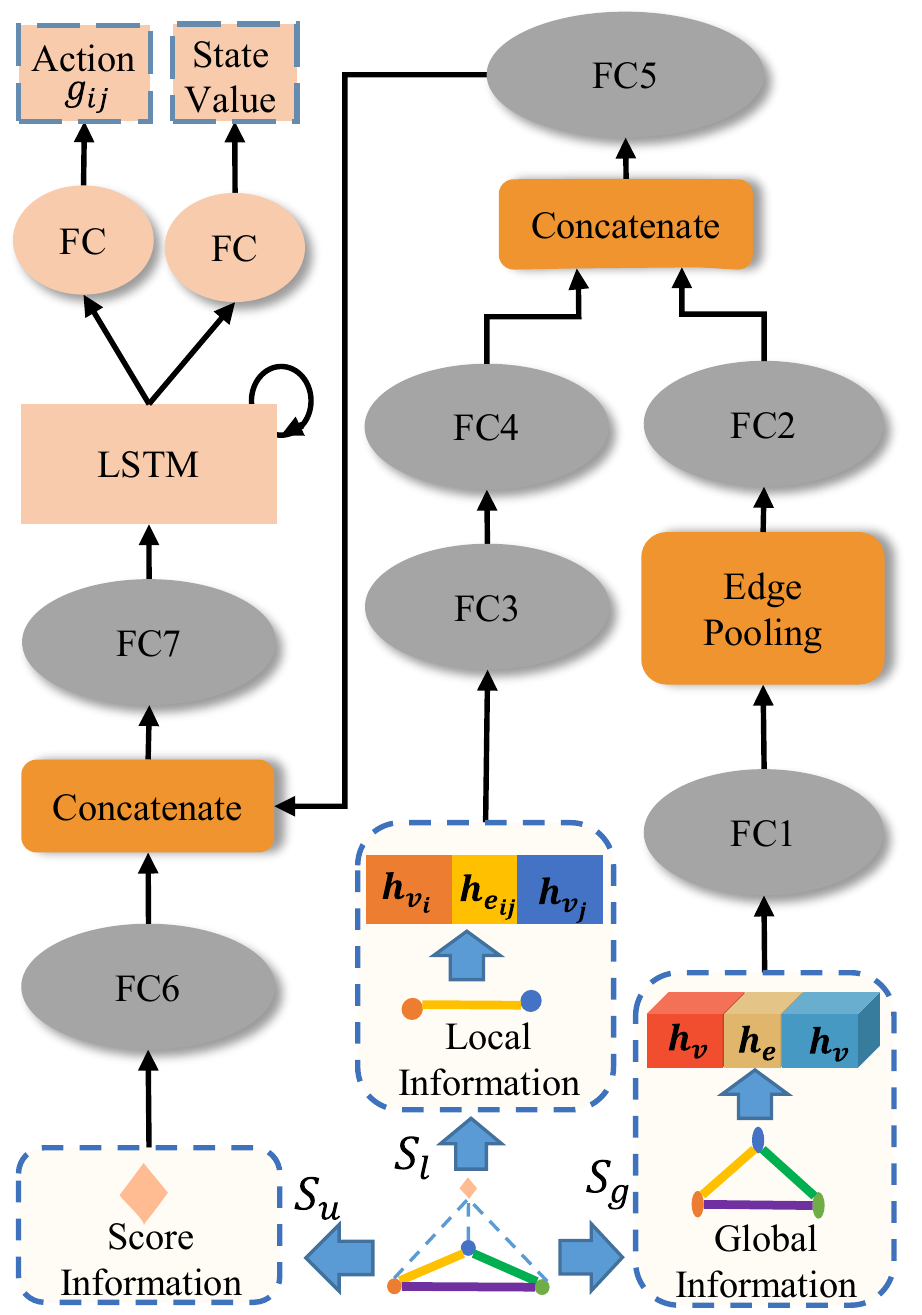}
		\caption{Structure of the RG agent}
		\label{RG:b}
	\end{subfigure}
	\label{RG}
	\caption{(a) Comparing the $ L_1 $ and $ L_{2,1} $ norms of gating matrix $ \bm{G} $, where the transparency denotes the value of each gate. The $ \|\bm{G}\|_1 $ encourages uniform sparsity while $ \|\bm{G}\|_{2,1} $ encourages structured row sparsity. The implementation of $ \|\bm{G}\|_{2,1} $ is illustrated in the bottom. (b) The RG agent takes in the global information $ S_g $, the local information $ S_l $ for specific relation, and the global scene attribute $ S_u $. ``FC1", \dots, ``FC7" are %
		fully connected layers, and ``Edge Pooling" denotes average pooling along the edge dimension. Finally, the left branch (\textit{Actor}) and the right branch (\textit{Critic}) outputs an action %
		and a value for the current state, respectively.
	}
\end{figure}

\textbf{Reward.} The reward $ r(S, A) $, reflecting the efficacy of action $ A $ w.r.t the state $ S $, consists of three parts. \text{1)} To encourage the relation gates $ \bm{G}=\{g_{ij}\}_{i,j=1}^{N_v} $ to selects group-relevant relations, we propose a structured sparsity reward. We define structured sparsity as the $ L_{2,1} $ norm of $ \bm{G} $, \ie, 
\begin{equation}
	L_{2,1}(\bm{G})= \sum_{i=1}^{N_v}\|\bm{g}_{i,:}\|_2 = \sum_{i=1}^{N_v}\left(\sqrt{\sum_{j=1}^{N_v} |g_{ij}|^2}\right) 
\end{equation}
where $ \bm{g}_{i,:} $ is row vectors of $ \bm{G} $. As illustrated in Fig.~\ref{RG:a}, unlike $ L_1$ norm that tends to uniformly make all gating elements sparse, the $ L_{2,1}$ norm can encourage the rows of $ \bm{G} $ to be sparse. Thus, the structured sparsity is very helpful to attend to a few key participants which have wide influence to others. The structured sparsity reward at the $ \tau $th reinforcement step is defined to encourage the agent to gradually attend to a few key participants and relations, \ie,
\begin{equation}
	r_{sparse} = - sgn\left( L_{2,1}\left( \bm{G}_{\tau}\right) - L_{2,1}\left(\bm{G}_{\tau-1}\right) \right) 
\end{equation}
where $ r_{sparse} \in \{-1,1\}$ and the $ sgn $ is sign function. \text{2)} To encourage the posterior probability to evolve along an ascending trajectory, we introduce an ascending reward with respect to the probability of groundtruth activity label, \ie, 
\begin{equation}
	r_{ascend} = sgn\left( \bm{p}^c_{\tau} - \bm{p}^c_{\tau-1} \right) 
\end{equation}
where $ \bm{p}^c_{\tau} $ is predicted probability of the groundtruth %
label at the $ \tau $th step. $ r_{ascend} \in \{-1,1\}$ reflects the probability improvement of the groundtruth.
\text{3)} To ensure that the model tends to predict correct classes, inspired by \cite{DBLP:conf/cvpr/TangTLL018}, a strong stimulation $ \Omega $ is enforced when
the predicted class shifts from wrong to correct after a step, and a strong punishment $ - \Omega $ is applied if the turning goes otherwise, \ie,
\begin{equation}
	r_{s} = 
	\begin{cases}
		\Omega, \qquad & \text{if stimulation}\\
		-\Omega, \qquad & \text{if punishment}\\
		0, \qquad & \text{otherwise}\\
	\end{cases}
	\label{r-EI}
\end{equation}
Finally, the total reward for the RG agent is
\begin{equation}
	r=r_{sparse} + r_{ascend} + r_{shift}.
\end{equation}

\textbf{Relation-gating Agent.} Since searching high dimensional continuous action space is challenging for reinforcement learning, we compromise to let the agent output one gating value at a time and cycle through all edges within each reinforcement step.
The architecture of the RG agent is shown in Fig.~\ref{RG:b}, which is under an Actor-Critic framework \cite{DBLP:conf/nips/KondaT99}. Inspired by human's decision making that historical experience can assist the current decision, a LSTM block is used to memorize the information of the past states. The agent maintains both a policy $ \pi(A_{\tau}|S_{\tau}; \theta) $ (also named \textit{Actor}) to generate actions (gates) and an estimation of value function $ V(S_{\tau};\theta_v) $ (also named \textit{Critic}) to assess values for corresponding states. Specifically, the \textit{Actor} outputs a mean $ \mu_{ij} $ and a standard deviation $ \sigma_{ij} $ of action distribution $ \mathcal{N}(\mu_{ij}, \sigma_{ij}) $. The action $ g_{ij} $ is sampled from the Gaussian distribution $ \mathcal{N}(\mu_{ij}, \sigma_{ij}) $ during training, and is set as $ \mu_{ij} $ directly during testing.

\textbf{Optimization.} The agent is optimized with the classical A3C algorithm \cite{DBLP:conf/icml/MnihBMGLHSK16} for reinforcement learning. The policy and the value function of the agent are updated after every $ \tau_{max} $ (updating interval) steps or when a terminal state is reached. The accumulated reward at the step $ \tau $ is {\mymathfont $ R_{\tau}=\sum_{i=0}^{k-1}\gamma^i r_{\tau+i}+\gamma^kV(S_{\tau+k};\theta_v) $}, where $ \gamma $ is the discount factor, $ r_{\tau} $  is the reward at the $ \tau th$ step, and $ k $ varies from 0 to $ \tau_{max} $. The advantage function can be calculated by {\mymathfont $ R_{\tau}-V(S_{\tau};\theta_v)$}, and the entropy of policy $ \pi $ is $H(\pi(S_{\tau}; \theta)) $. Eventually, the gradients are accumulated via Eq.~\ref{eq_critic} and Eq.~\ref{eq_actor} to respectively update the value function and the policy of agent \cite{DBLP:conf/icml/MnihBMGLHSK16}.

{\mymathfont
	\begin{equation}
		d\theta_v \leftarrow d\theta_v + \nabla_{\theta_v}\left(R_{\tau}-V(S_{\tau};\theta_v)\right)^2/2 
		\label{eq_critic}
	\end{equation}
	\begin{equation}
		\begin{split}
			d\theta \leftarrow d\theta &+ \nabla_{\theta}log\pi(A_\tau|S_\tau;\theta)\left(R_{\tau}-V(S_{\tau};\theta_v)\right)\\
			&+\beta \nabla_{\theta} H(\pi(S_{\tau}; \theta)) 
		\end{split}
		\label{eq_actor}
	\end{equation}
}where $ \beta $ controls the strength of entropy regularization.

\begin{figure}[tbp]
	\centering
	\begin{subfigure}[t]{0.95\linewidth}
		\includegraphics[width=\linewidth,height=0.95\linewidth]{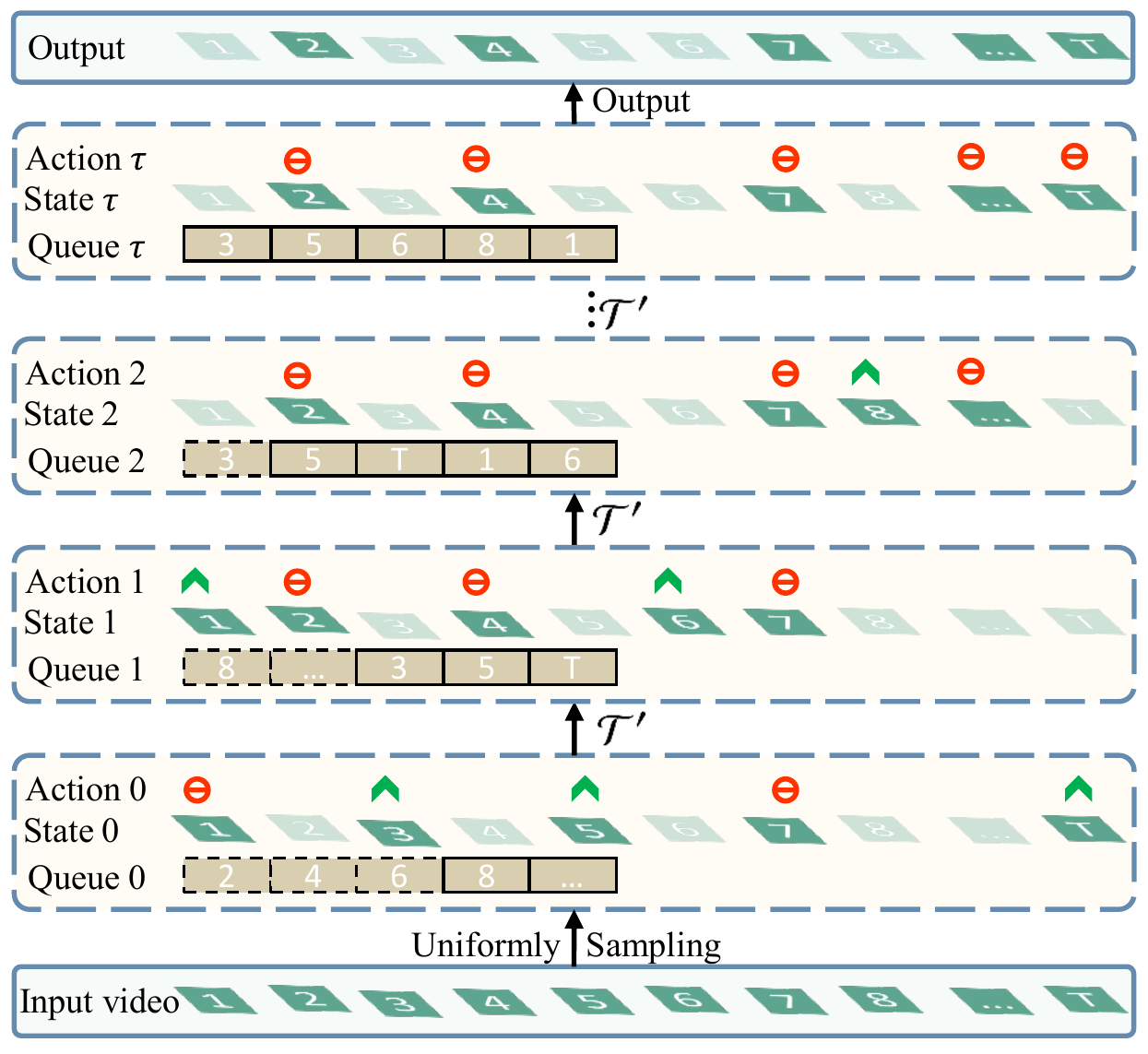}
		\caption{Illustration of the feature-distilling process}
		\label{FA:a}
	\end{subfigure}
	\begin{subfigure}[t]{\linewidth}
		\includegraphics[width=\linewidth,height=0.45\linewidth]{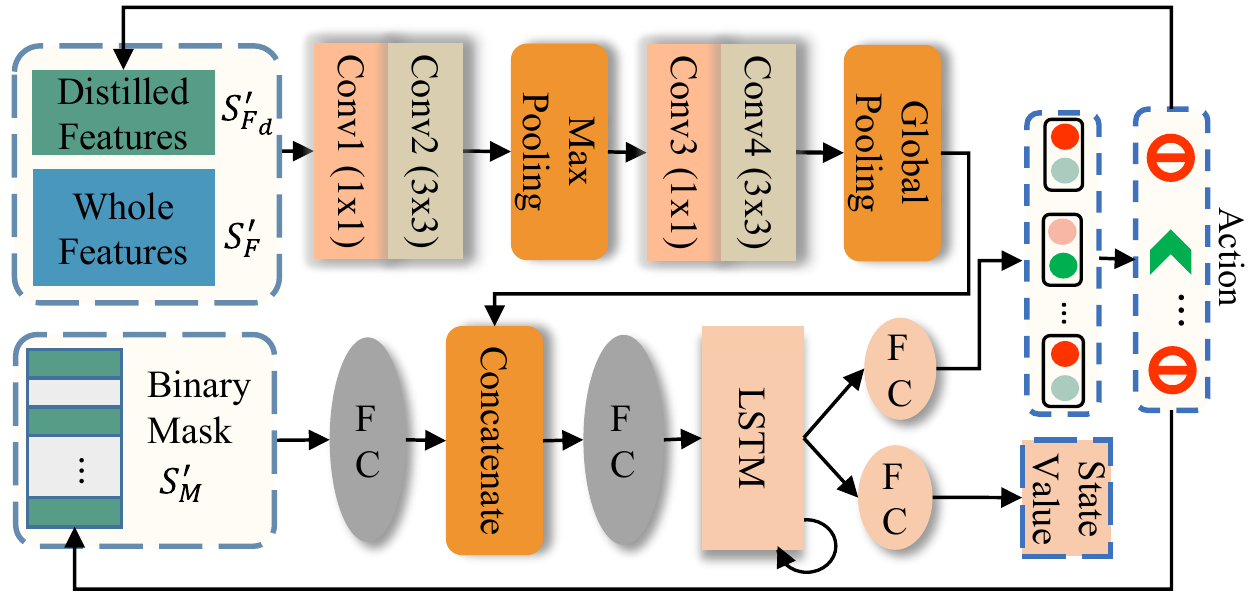}
		\caption{Structure of the feature-distilling agent}
		\label{FA:b}
	\end{subfigure}
	\caption{(a) The FD agent has two discrete actions, \ie, ``\textit{stay distilled}" (red icon) and ``\textit{shift to alternate}" (green icon). The ``Queue" is a queue which contains the alternate feature frames, and $ \mathcal{T}' $ is the deterministic state transition function. %
		(b) %
		The convolutional layers Conv1 and Conv3 (with kernel of 1x1) are used for channel squeezing, and Conv2 and Conv4 (with kernel of 3x3) are used for feature extracting. The``FC" denotes fully connected layer.
	}
	\label{FA}
\end{figure}

\subsection{Progressively Feature Distilling}
To further refine the low-level spatiotemporal features used for constructing %
graph, we introduced another feature-distilling agent. It is aimed at distilling the most informative frames of features, which is also formulated as
a Markov Decision Process {\mymathfont$ \mathcal{M}'=\{S',A', \mathcal{T}', r', \gamma' \}$}.

\textbf{State.} The state of the FD agent consists of three components $S'=\{S'_F,S'_{F_d}, S'_M\}$. The whole feature tensor of an activity $ S'_F \in \mathbb{R}^{N\times T\times D_F} $ provides the global information about the activity clip, 
where $ N $, $ T $ and $ D_F $ are respectively the numbers of person, frame and feature dimension of the %
feature tensor. The local feature $ S'_{F_d} \in \mathbb{R}^{N\times T_d\times D_F} $ carries the \textit{implicit} information of the distilled frames, where $ T_d $ is the number of frames to be kept. In order to be \textit{explicitly} aware of the distilled frames, the state of FD agent also contains the binary mask $ S'_M  $ of the distilled frames.

\textbf{Action.} As shown in Fig.~\ref{FA:a}, the FD agent outputs two types of discrete actions for each selected frame, \ie~``\textit{stay distilled}" indicating the frame is informative that the agent determines to keep it, and ``\textit{shift to alternate}" indicating the agent determines to discard the frame and take in an alternate. The shifting may be frequent at the beginning but will gradually become stable after some explorations (Fig.~\ref{FA:a}). In order to give equal chance for all alternates to be enrolled, the latest discarded frames are appended to the end of a queue and have the lowest priority to be enrolled again.

\textbf{Feature-distilling Agent.} %
The FD agent in Fig.~\ref{FA:b} is also constructed under the Actor-Critic \cite{DBLP:conf/nips/KondaT99} framework. The agent takes in the global knowledge from the whole feature {\mymathfont$ S'_F $}, the \textit{implicit} local knowledge from the distilled features {\mymathfont$ S'_{F_d} $}, and the \textit{explicit} local knowledge from the binary frame mask {\mymathfont$ S'_M $}. Finally, the agent outputs an action vector for the $ T_d $ distilled feature frames and a value for the current state. The action vector is sampled from the policy distribution during training, and is directly set as the action type with max probability during testing.

\textbf{Optimization and Rewards.} The optimization algorithm (A3C) and object function are same as the RG agent. %
The reward only contains the components about trajectory ascending  and class shifting introduced above, \ie, 
\begin{equation}
	r'= r_{ascend} + r_{shift}.
\end{equation}

\subsection{Training Procedure}
In the proposed approach, the agents and the graph need to be updated respectively on CPU (to exploit numerous CPU cores/threads for asynchronous updating workers according to A3C algorithm \cite{DBLP:conf/icml/MnihBMGLHSK16}) and GPU. In addition, the graph is updated after each video batch, but the agents are updated many times during each video when the number of reinforcement step reaches the updating interval $ \tau_{max} $ or a terminal state is reached. Thus, the graph and agents are updated on different devices with different updating periods, and it is unable to optimize them with conventional end-to-end training. Therefore, we adopt alternate training. More details of the standard flowchart of A3C algorithm can be found in the \textit{Supplementary Material}.

\textbf{Individual Feature Preparation.} Following \cite{tang2018mining}, we finetune the \text{Person-CNN} (VGG16 \cite{DBLP:journals/corr/SimonyanZ14a}) 
pretrained on ImageNet \cite{DBLP:journals/ijcv/RussakovskyDSKS15} with individual action labels to extract visual features, and then train the \text{Person-LSTM} with individual action labels to extract temporal features. To lower the computation burden, the extracted individual features are saved to disk and only need reloading after this procedure. 

\textbf{Alternate Training.} There are totally 9 separated training stages. At each stage, only one of the three components (SRG, trained with 15 epochs; FD- or RG-agent, trained with 2 hours) is trained and the remaining two are frozen (or removed). In the first stage, the SRG (without agents) is trained with the extract features to capture the context information within activities. In the second stage, the SRG is frozen, and the FD agent is introduced and trained with the rewards provided by the frozen SRG. In the third stage, the SRG and FD agent are frozen, the RG agent is introduced and trained with the rewards provided by the frozen SRG and FD agent. After that, one of the SRG, FD agent and RG agent is trained in turn with the remaining two be frozen in the following 6 stages.

\section{Experiments}
\subsection{Datasets}
\textbf{Volleyball Datasets} \cite{DBLP:conf/cvpr/IbrahimMDVM16}. The Volleyball dataset is currently the largest dataset for group activity recognition. It contains 4830 clips of 55 volleyball videos. %
Each clip is annotated with 8 group activity categories (\ie, right set, right spike, right pass, right winpoint, left winpoint, left pass, left spike and left set), and its middle frame is annotated with 9 individual action labels (\ie, waiting, setting, digging, falling, spiking, blocking, jumping, moving and standing). %
We %
employ the metrics of Multi-class Classification Accuracy (MCA) and Mean Per Class Accuracy (MPCA) to evaluate the performance %
following \cite{yan2018participation}.

\textbf{Collective Activity Dataset (CAD)} \cite{choi2009they}. The CAD contains 2481 activity clips of 44 videos. %
The middle frame of each clip is annotated with 6 individual action classes (\ie, NA, crossing, walking, waiting, talking and queueing), and the group activity label is assigned as the majority action label of individuals in the scene. %
Following \cite{DBLP:conf/cvpr/WangNY17}, we merge the classes ``walking" and ``crossing" as ``moving" and report the %
MPCA to evaluate the performance.

Since the existing datasets lack sufficient diversity of background \cite{DBLP:conf/cvpr/WangNY17}, it is too difficult to distinguish useful objects (\eg, volleyball) from noisy background without any annotation. Following \cite{DBLP:conf/cvpr/IbrahimMDVM16,DBLP:conf/iccv/LiC17,yan2018participation,DBLP:conf/cvpr/ShuTZ17,DBLP:conf/eccv/QiQLWLG18,tang2018mining}, we ignore the background and only focus on interactions among persons.

\subsection{Implementation Details}
For fair comparison with previous methods \cite{DBLP:conf/eccv/QiQLWLG18,tang2018mining}, we use the same backbone network (Person-CNN) VGG16 \cite{DBLP:journals/corr/SimonyanZ14a}. It outputs 4096-d features and the Person-LSTM equipped with 3000 hidden neurons takes in all the features in T (T=10) time steps. In the SRG, the embedding sizes of node and edge are 1000 and 100 respectively, and the graph passes 3 iterations at each time. Thus, the number of hidden neurons in updating functions $ \text{NN}_{ve} $, $ \text{NN}_v $, and  $ \text{NN}_e $ are 1000, 1000 and 100, respectively. In the RG agent, the fully connected layers FC1, FC2, \dots, FC7 are respectively contains 512, 256, 512, 256, 256, 64 and 256 neurons, and its LSTM network contains 128 hidden nodes. In the FD agent, the number of feature frames to be kept $ T_d $ is practically set as 5. In Fig.~\ref{FA:b}, the neuron number of the two FC layers from the left to right is 64 and 256, the channels of Conv1, Conv2, Conv3, Conv4 are respectively 1024, 1024, 256, 256, and the LSTM network contains 128 neurons. 

During training, we use RMSprop/Adam (SRG/Agents) optimizer with an initial learning rate of 0.00001/0.0001 (SRG/Agents) and a weight decay of 0.0001. The batch size is 8/16 (CAD/Volleyball) for SRG training. The discount factor $ \gamma $, entropy factor $ \beta $ and the number of asynchronous workers in A3C for both agents are respectively set as 0.99, 0.01 and 16. In practice, the updating interval $ \tau_{max} $ and $ \Omega $ (in Eq.~\ref{r-EI}) are set as 5/5 and 15/20 (RG/FD agent), respectively. In Volleyball dataset, following %
\cite{DBLP:conf/cvpr/IbrahimMDVM16}, the 12 players are split into two subgroups (\ie, the left team and the right team) according to %
positions, and the RG agent are shared by the two subgroups in our framework, and finally the outputs of the two subgroups are averaged. In CAD dataset, since the number of individuals is varying from 1 to 12, we select 5 effective persons for each frame and fill zeros for the frames contain less than 5 persons following %
\cite{yan2018participation}.

\subsection{Baseline and Variants for Ablation Studies}
To examine the effectiveness of each component in the proposed method, we conduct ablation studies with the following baseline and variants. \textit{stagNet w/o Atten.}~\cite{DBLP:conf/eccv/QiQLWLG18}: this baseline constructs a message passing graph network with the similar low-level features as our SRG. It implicitly represents the interactions by the passing messages, while our SRG explicitly models relations in a full graph network. \textit{Ours-SRG}: this variant only contains the SRG of the proposed method. \textit{Ours-SRG+T.~A.}: this variant contains our SRG and a temporal attention over feature frames. \textit{Ours-SRG+R.~A.}: this variant contains our SRG and a relation attention that directly learns relation gates. \textit{Ours-SRG+FD}: this variant contains both the SRG and FD agent, and they are trained alternately to boost each other. \textit{Ours-SRG+RG}: this variant contains both the SRG and RG agent, and they are alternately trained. \textit{Ours-SRG+FD+RG (PRL)}: our progressive reinforcement learning framework that contains all the proposed three components, including the SRG, the FD agent, and the RG agent.

\begin{table}[tbp]
	\caption{Comparisons of recognition accuracy (\%) on Volleyball dataset. ``OF" denotes additional optical flow input.}
	\label{tab-volleybal}
	\begin{tabular}{lcccc}
		\toprule
		Methods&Backbone&OF&MCA &MPCA\\
		\midrule
		HDTM~\cite{DBLP:conf/cvpr/IbrahimMDVM16}&AlexNet&N&81.9&82.9\\
		SBGAR~\cite{DBLP:conf/iccv/LiC17}&{\small Inception-v3}&Y&66.9&67.6\\
		CERN-2~\cite{DBLP:conf/cvpr/ShuTZ17}&VGG16&N&83.3&83.6\\
		SSU~\cite{DBLP:conf/cvpr/BagautdinovAFFS17}&{\small Inception-v3}&N&89.9&-\\
		SRNN~\cite{DBLP:conf/wacv/BiswasG18}&AlexNet&N&83.5&-\\
		PC-TDM~\cite{yan2018participation}&AlexNet&Y&87.7&88.1\\
		stagNet~\cite{DBLP:conf/eccv/QiQLWLG18}&VGG16&N&89.3&-\\
		{\small SPA+KD}~\cite{tang2018mining}&VGG16&N&89.3&89.0\\
		{\small SPA+KD+OF}~\cite{tang2018mining}&VGG16&Y&90.7&90.0\\
		ARG~\cite{wu2019learning}&VGG16&N&\textbf{91.9}&-\\
		CRM~\cite{azar2019convolutional}&I3D&Y&\textbf{93.0}&-\\
		\midrule
		Baseline \cite{DBLP:conf/eccv/QiQLWLG18}&VGG16&N&87.9&-\\
		Ours-SRG&VGG16&N&88.3&88.5\\	
		Ours-SRG+T. A.&VGG16 &N&88.6&88.7\\
		Ours-SRG+R. A.&VGG16 &N&88.7&89.0\\	
		Ours-SRG+FD&VGG16&N&89.5&89.2\\
		Ours-SRG+RG&VGG16&N&89.8&91.1\\	
		Ours-PRL&VGG16&N&\textbf{91.4}&\textbf{91.8}\\			
		\bottomrule
	\end{tabular}
\end{table}

\begin{figure}[tbp]
	\centering
	\includegraphics[width=0.94\linewidth,height=0.89\linewidth]{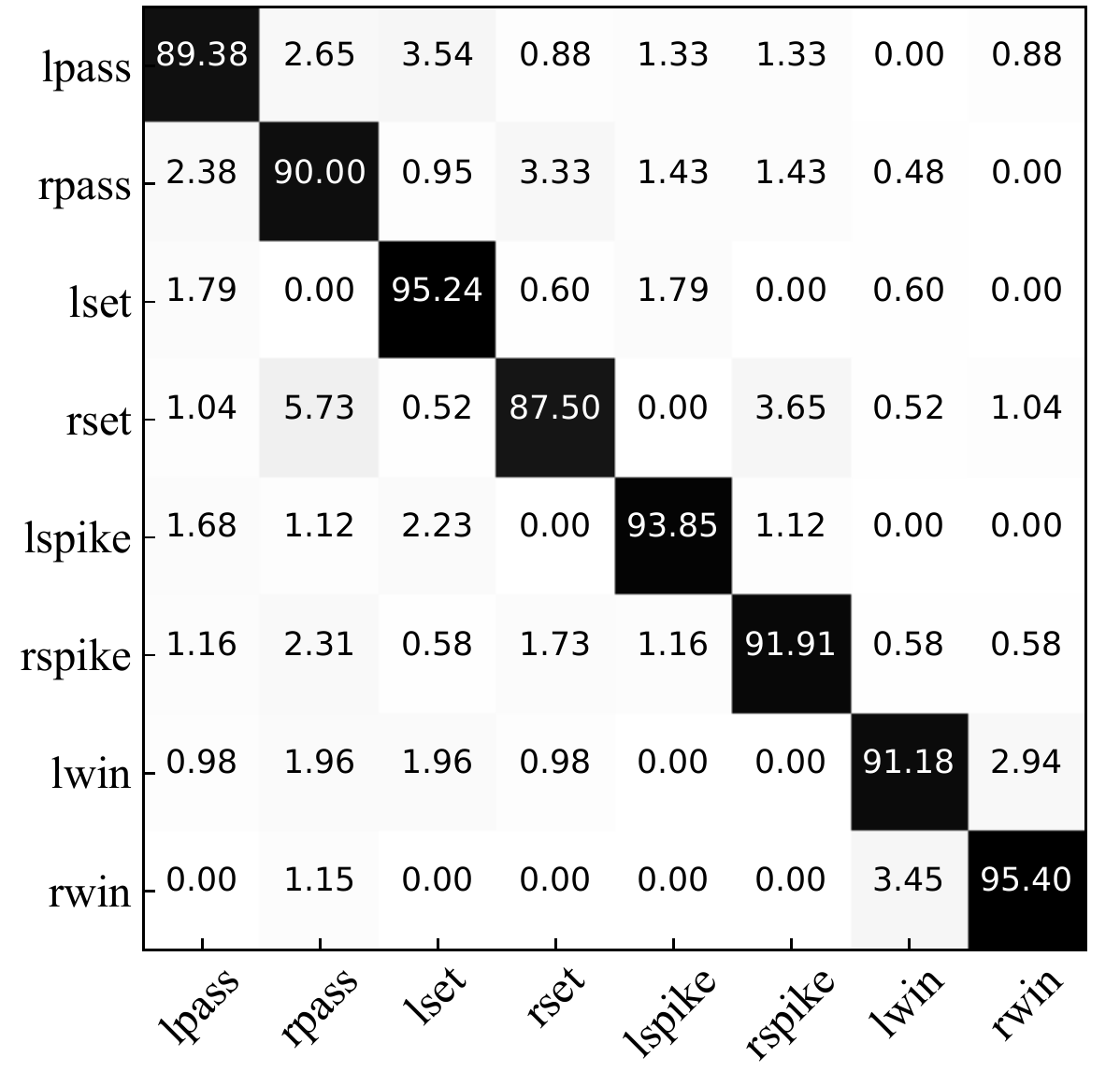}
	\caption{Confusion matrix on the Volleyball dataset.}
	\label{CM_Volleyball}
\end{figure}

\subsection{Results on the Volleyball Dataset}
To examine the effectiveness of each component, we compare the proposed PRL against the above baseline and variants. As Table~\ref{tab-volleybal} shows, although building graphs on similar low-level features, our semantic relation graph is superior to the baseline (\text{stagNet w/o Atten.} \cite{DBLP:conf/eccv/QiQLWLG18})
because our semantic relations are explicitly modeled while the baseline only implicitly contains them in the passing messages. Our SRG+FD boosts the SRG over 1.2\% (MCA) and 0.7\% (MPCA) by applying the FD agent to filter out ambiguous frames of features, and our SRG+RG also improves the performance of the SRG over 1.5\% (MCA) and 2.6\% (MPCA) by exploiting the RG agent to refine the relations. Our PRL achieves better performance by combining the advantages from the two agents. Note that the PRL eventually improves 3.1\% (MCA) over the original SRG, which is even larger than the sum of increments from the two agents, 2.7\% (MCA), indicating that the two agents can boost each other through the alternate training procedure. Besides, the agent-equipped variants SRG+FD and SRG+RG respectively perform better than corresponding attention-equipped variants \text{SRG+T.~A.} and \text{SRG+R.~A.} by 0.9\% and 1.1\% (MCA). The superiority of the agents probably owe to two reasons: 1) The attention variants can only learn from the annotated activity labels, while our RL-based agents can also learn from the historical experience during the policy exploring processes. 2) The attention variants only updates for each video batch, while our agents are updated many times during each single video (cf.~training flowchart) that can achieve more fine-grained and video-specific adjustments. 

Then, we compare the proposed PRL with other state-of-the-art methods. As shown in Table~\ref{tab-volleybal}, our PRL is on par with the state-of-the-art method that has no extra optical flow input (ARG \cite{wu2019learning}). Our PRL even outperforms most of the methods that exploit optical flow input (including SBGAR \cite{DBLP:conf/iccv/LiC17}, PC-TDM \cite{yan2018participation}, and SPA+KD+OF \cite{tang2018mining}). Although CRM \cite{azar2019convolutional} performs somewhat better than our PRL, it is unfair to compare with. Because the CRM not only exploits extra optical flow input but only utilizes a much larger backbone (I3D~\cite{carreira2017quo}) than ours (VGG16 \cite{DBLP:journals/corr/SimonyanZ14a}).

\begin{figure*}[tbp]
	\centering
	\includegraphics[width=1.0\linewidth,height=0.39\linewidth]{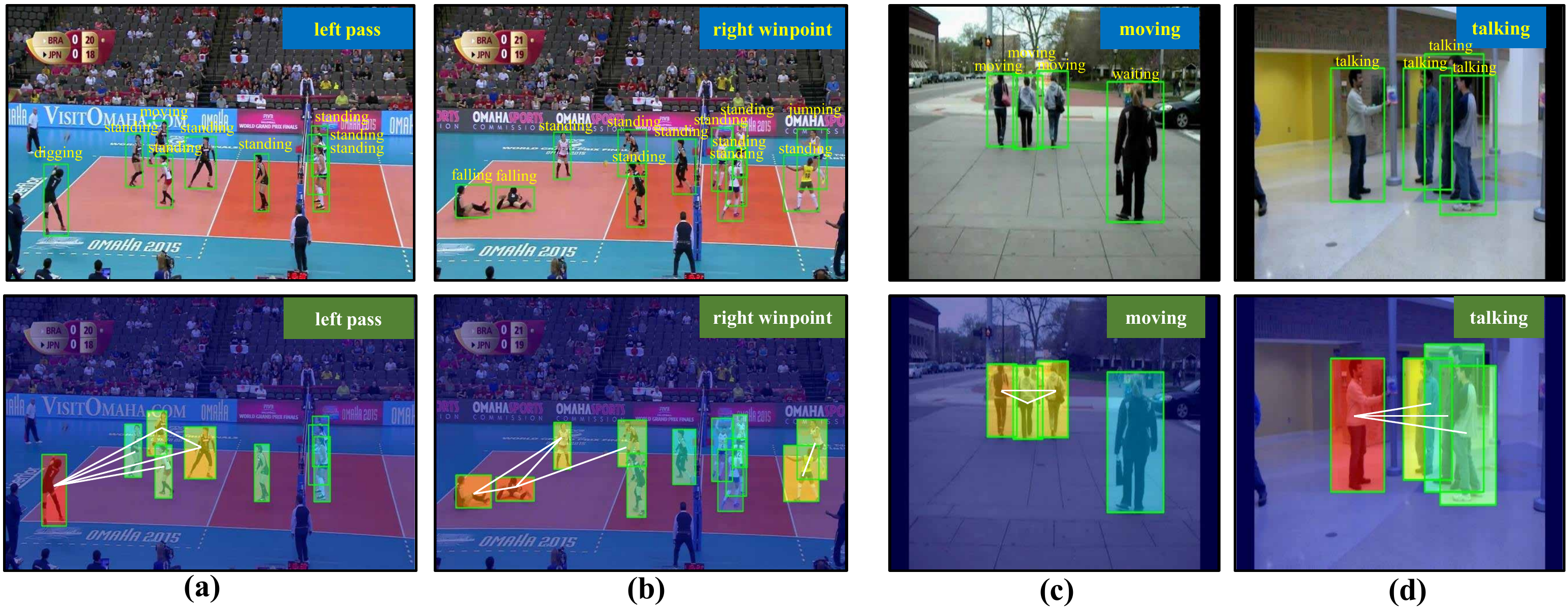}
	\caption{Visualization of the refined SRGs. The first row contains the obtained tracklets and the groundtruth labels of activity and person actions. The second row contains the refined SRGs and the predicted activity labels. The color of person represents its importance degree. To facilitate visualization, only the relations with top5/top3 (Volleyball/CAD) gate values are shown (the white lines). The samples of (a,b) and (c,d) are from the Volleyball and CAD datasets, respectively.}
	\label{visualization}
\end{figure*}

In addition, the confusion matrix of the proposed PRL is shown in Fig.~\ref{CM_Volleyball}. As we can see, our PRL achieves promising recognition accuracies ($ \geq 90\%$) on most of the activities. The main failure cases are from ``set" and ``pass" within the left and right subgroups, which is probably due to the very similar actions and positions of the key participants. We also visualized several refined semantic relation graphs in Fig.~\ref{visualization}, where the relations with top5 gate values are shown and the importance degree of persons are indirectly computed by summing the connected relation gates (normalized over all persons). In Fig.~\ref{visualization}{\color{red}a}, benefited from the rewards of structured
sparsity, our RG agent successfully discovers the subset of relations related to the ``digging" person is the key to determine the activity ``left pass". In Fig.~\ref{visualization}{\color{red}b}, the model predicts ``right winpoint" mainly based on two relation clusters, including the cluster characterized by the two ``falling" persons in the left team and the cheering cluster in the right team.

\begin{table}
	\caption{Comparisons of recognition accuracy (\%) on CAD dataset. ``OF" denotes additional optical flow input.}
	\label{tab-cad}
	\centering
	\begin{threeparttable}[htb]
		\begin{tabular}{lccc}
			\toprule
			Methods&Backbone&OF&MPCA(\%)\\
			\midrule
			HDTM \cite{DBLP:conf/cvpr/IbrahimMDVM16}&AlexNet&N&89.6\\
			CERN-2 \cite{DBLP:conf/cvpr/ShuTZ17}&VGG16&N&88.3\\
			SBGAR \cite{DBLP:conf/iccv/LiC17}&Inception-v3&Y&89.9\\
			PC-TDM \cite{yan2018participation}&AlexNet&Y&92.2\\
			SPA+KD \cite{tang2018mining}&VGG16&N&92.5\\
			SPA+KD+OF \cite{tang2018mining}&VGG16&Y&\textbf{95.7}\\
			CRM \cite{azar2019convolutional}&I3D&Y&94.2\\
			\midrule
			Baseline \cite{DBLP:conf/eccv/QiQLWLG18}&VGG16&N&87.7\tnote{\text{*}}\\
			Ours-SRG &VGG16&N&89.4\\
			Ours-SRG+R.~A. &VGG16&N&90.0\\
			Ours-SRG+T.~A. &VGG16&N&90.1\\	
			Ours-SRG+FD&VGG16&N&91.1\\
			Ours-SRG+RG&VGG16&N&91.4\\	
			Ours-PRL &VGG16&N&\textbf{93.8}\\
			\bottomrule
		\end{tabular}
		\begin{tablenotes}
			\item[\text{*}] MPCA is unavailable, MCA is listed instead.
		\end{tablenotes}
	\end{threeparttable}
\end{table}

\subsection{Results on the Collective Activity Dataset}

Table~\ref{tab-cad} shows the comparison with different methods on the CAD dataset. Following \cite{yan2018participation,tang2018mining}, the results regarding MPCA of several methods are calculated from the reported confusion matrices in \cite{DBLP:conf/cvpr/HajimirsadeghiY15,DBLP:conf/cvpr/IbrahimMDVM16,DBLP:conf/cvpr/ShuTZ17,DBLP:conf/iccv/LiC17}. %
Our PRL outperforms the state-of-the-art method (SPA+KD \cite{tang2018mining}) without extra optical flow input by a margin of 1.3\%. Although the SPA+KD+OF \cite{tang2018mining} performs better than our PRL, its main improvement (3.2\%) is owed to the extra optical flow information (cf.~Table~\ref{tab-cad}). The backbone of CRM~\cite{azar2019convolutional} (I3D) is much larger than ours (VGG19), making it less comparable. The detailed confusion matrix of our PRL on the CAD dataset can also be found in the \textit{Supplementary Material}.

Furthermore, we analyze the results by visualizing the final SRGs. For the ``Moving" activity in Fig.~\ref{visualization}{\color{red}c}, our method concentrates on the relations among the three moving persons to suppress the noisy relations caused by the ``Waiting" person. Similarly, in Fig.~\ref{visualization}{\color{red}d}, our method successfully attends to the relations connected to the ``Talking" person and weakens the relations among the three audiences.

\section{Conclusion}
In this work, we propose a novel progressive relation learning method to model and distill the group-relevant actions and interactions in group activities. A graph built on the spatiotemporal features and the interactions of individuals is used to explicitly model the semantic relations in group activities. A feature-distilling agent is proposed to progressively distill the most informative frames of the low-level features, and the relation-gating agent is proposed to refine the high-level relations in the semantic relation graph. Eventually, our PRL achieves promising results on two widely used benchmarks for group activity recognition.

{\small
\bibliographystyle{ieee_fullname}
\bibliography{egbib}

\begin{thebibliography}{10}\itemsep=-1pt

\bibitem{azar2019convolutional}
Sina~Mokhtarzadeh Azar, Mina~Ghadimi Atigh, Ahmad Nickabadi, and Alexandre
  Alahi.
\newblock Convolutional relational machine for group activity recognition.
\newblock In {\em CVPR}, pages 7892--7901, 2019.

\bibitem{DBLP:conf/cvpr/BagautdinovAFFS17}
Timur~M. Bagautdinov, Alexandre Alahi, Fran{\c{c}}ois Fleuret, Pascal Fua, and
  Silvio Savarese.
\newblock Social scene understanding: End-to-end multi-person action
  localization and collective activity recognition.
\newblock In {\em CVPR}, pages 3425--3434, 2017.

\bibitem{DBLP:journals/corr/abs-1806-01261}
Peter~W. Battaglia, Jessica~B. Hamrick, Victor Bapst, Alvaro
  Sanchez{-}Gonzalez, Vin{\'{\i}}cius~Flores Zambaldi, Mateusz Malinowski,
  Andrea Tacchetti, David Raposo, Adam Santoro, Ryan Faulkner, {\c{C}}aglar
  G{\"{u}}l{\c{c}}ehre, Francis Song, Andrew~J. Ballard, Justin Gilmer,
  George~E. Dahl, Ashish Vaswani, Kelsey Allen, Charles Nash, Victoria
  Langston, Chris Dyer, Nicolas Heess, Daan Wierstra, Pushmeet Kohli, Matthew
  Botvinick, Oriol Vinyals, Yujia Li, and Razvan Pascanu.
\newblock Relational inductive biases, deep learning, and graph networks.
\newblock {\em arXiv preprint arXiv:1806.01261}, 2018.

\bibitem{DBLP:conf/wacv/BiswasG18}
Sovan Biswas and Juergen Gall.
\newblock Structural recurrent neural network {(SRNN)} for group activity
  analysis.
\newblock In {\em WACV}, pages 1625--1632, 2018.

\bibitem{DBLP:journals/spm/BronsteinBLSV17}
Michael~M. Bronstein, Joan Bruna, Yann LeCun, Arthur Szlam, and Pierre
  Vandergheynst.
\newblock Geometric deep learning: Going beyond euclidean data.
\newblock {\em {IEEE} Signal Process. Mag.}, 34(4):18--42, 2017.

\bibitem{carreira2017quo}
Joao Carreira and Andrew Zisserman.
\newblock Quo vadis, action recognition? a new model and the kinetics dataset.
\newblock In {\em CVPR}, pages 6299--6308, 2017.

\bibitem{choi2012unified}
Wongun Choi and Silvio Savarese.
\newblock A unified framework for multi-target tracking and collective activity
  recognition.
\newblock In {\em ECCV}, pages 215--230. Springer, 2012.

\bibitem{choi2009they}
Wongun Choi, Khuram Shahid, and Silvio Savarese.
\newblock What are they doing?: Collective activity classification using
  spatio-temporal relationship among people.
\newblock In {\em ICCV Workshops}, pages 1282--1289. IEEE, 2009.

\bibitem{DBLP:conf/cvpr/DengVHM16}
Zhiwei Deng, Arash Vahdat, Hexiang Hu, and Greg Mori.
\newblock Structure inference machines: Recurrent neural networks for analyzing
  relations in group activity recognition.
\newblock In {\em CVPR}, pages 4772--4781, 2016.

\bibitem{DBLP:conf/icml/GilmerSRVD17}
Justin Gilmer, Samuel~S. Schoenholz, Patrick~F. Riley, Oriol Vinyals, and
  George~E. Dahl.
\newblock Neural message passing for quantum chemistry.
\newblock In {\em ICML}, pages 1263--1272, 2017.

\bibitem{DBLP:conf/cvpr/HajimirsadeghiY15}
Hossein Hajimirsadeghi, Wang Yan, Arash Vahdat, and Greg Mori.
\newblock Visual recognition by counting instances: {A} multi-instance
  cardinality potential kernel.
\newblock In {\em CVPR}, pages 2596--2605, 2015.

\bibitem{hu2019joint}
Guyue Hu, Bo Cui, and Shan Yu.
\newblock Joint learning in the spatio-temporal and frequency domains for
  skeleton-based action recognition.
\newblock {\em IEEE Transactions on Multimedia}, 2019.

\bibitem{DBLP:journals/icme/gyhu}
Guyue Hu, Bo Cui, and Shan Yu.
\newblock Skeleton-based action recognition with synchronous local and
  non-local spatio-temporal learning and frequency attention.
\newblock In {\em 2019 IEEE International Conference on Multimedia and Expo
  (ICME)}, pages 1216--1221, 2019.

\bibitem{DBLP:conf/cvpr/IbrahimMDVM16}
Mostafa~S. Ibrahim, Srikanth Muralidharan, Zhiwei Deng, Arash Vahdat, and Greg
  Mori.
\newblock A hierarchical deep temporal model for group activity recognition.
\newblock In {\em CVPR}, pages 1971--1980, 2016.

\bibitem{DBLP:journals/jmlr/King09}
Davis~E. King.
\newblock Dlib-ml: {A} machine learning toolkit.
\newblock {\em Journal of Machine Learning Research}, 10:1755--1758, 2009.

\bibitem{DBLP:conf/nips/KondaT99}
Vijay~R. Konda and John~N. Tsitsiklis.
\newblock Actor-critic algorithms.
\newblock In {\em NIPS}, pages 1008--1014, 1999.

\bibitem{DBLP:journals/pami/LanWYRM12}
Tian Lan, Yang Wang, Weilong Yang, Stephen~N. Robinovitch, and Greg Mori.
\newblock Discriminative latent models for recognizing contextual group
  activities.
\newblock {\em {IEEE} Trans. Pattern Anal. Mach. Intell.}, 34(8):1549--1562,
  2012.

\bibitem{DBLP:conf/cvpr/LiWZH18}
Debang Li, Huikai Wu, Junge Zhang, and Kaiqi Huang.
\newblock {A2-RL:} aesthetics aware reinforcement learning for image cropping.
\newblock In {\em CVPR}, pages 8193--8201, 2018.

\bibitem{DBLP:conf/iccv/LiC17}
Xin Li and Mooi~Choo Chuah.
\newblock {SBGAR:} semantics based group activity recognition.
\newblock In {\em ICCV}, pages 2895--2904, 2017.

\bibitem{DBLP:conf/icml/MnihBMGLHSK16}
Volodymyr Mnih, Adri{\`{a}}~Puigdom{\`{e}}nech Badia, Mehdi Mirza, Alex Graves,
  Timothy~P. Lillicrap, Tim Harley, David Silver, and Koray Kavukcuoglu.
\newblock Asynchronous methods for deep reinforcement learning.
\newblock In {\em ICML}, 2016.

\bibitem{DBLP:journals/corr/MnihKSGAWR13}
Volodymyr Mnih, Koray Kavukcuoglu, David Silver, Alex Graves, Ioannis
  Antonoglou, Daan Wierstra, and Martin~A. Riedmiller.
\newblock Playing atari with deep reinforcement learning.
\newblock {\em arXiv preprint arXiv:1312.5602}, 2013.

\bibitem{DBLP:conf/eccv/QiQLWLG18}
Mengshi Qi, Jie Qin, Annan Li, Yunhong Wang, Jiebo Luo, and Luc~Van Gool.
\newblock stagnet: An attentive semantic {RNN} for group activity recognition.
\newblock In {\em ECCV}, pages 104--120, 2018.

\bibitem{DBLP:journals/ijcv/RussakovskyDSKS15}
Olga Russakovsky, Jia Deng, Hao Su, Jonathan Krause, Sanjeev Satheesh, Sean Ma,
  Zhiheng Huang, Andrej Karpathy, Aditya Khosla, Michael~S. Bernstein,
  Alexander~C. Berg, and Fei{-}Fei Li.
\newblock Imagenet large scale visual recognition challenge.
\newblock {\em IJCV}, 115(3):211--252, 2015.

\bibitem{DBLP:conf/icml/Sanchez-Gonzalez18}
Alvaro Sanchez{-}Gonzalez, Nicolas Heess, Jost~Tobias Springenberg, Josh Merel,
  Martin~A. Riedmiller, Raia Hadsell, and Peter Battaglia.
\newblock Graph networks as learnable physics engines for inference and
  control.
\newblock In {\em ICML}, pages 4467--4476, 2018.

\bibitem{DBLP:conf/cvpr/ShuTZ17}
Tianmin Shu, Sinisa Todorovic, and Song{-}Chun Zhu.
\newblock {CERN:} confidence-energy recurrent network for group activity
  recognition.
\newblock In {\em CVPR}, pages 4255--4263, 2017.

\bibitem{DBLP:conf/cvpr/ShuXRTZ15}
Tianmin Shu, Dan Xie, Brandon Rothrock, Sinisa Todorovic, and Song{-}Chun Zhu.
\newblock Joint inference of groups, events and human roles in aerial videos.
\newblock In {\em CVPR}, pages 4576--4584, 2015.

\bibitem{DBLP:conf/cvpr/SimonovskyK17}
Martin Simonovsky and Nikos Komodakis.
\newblock Dynamic edge-conditioned filters in convolutional neural networks on
  graphs.
\newblock In {\em CVPR}, pages 29--38, 2017.

\bibitem{DBLP:journals/corr/SimonyanZ14a}
Karen Simonyan and Andrew Zisserman.
\newblock Very deep convolutional networks for large-scale image recognition.
\newblock In {\em ICLR}, 2015.

\bibitem{DBLP:conf/nips/SuttonMSM99}
Richard~S. Sutton, David~A. McAllester, Satinder~P. Singh, and Yishay Mansour.
\newblock Policy gradient methods for reinforcement learning with function
  approximation.
\newblock In {\em NIPS}, pages 1057--1063, 1999.

\bibitem{DBLP:conf/cvpr/TangTLL018}
Yansong Tang, Yi Tian, Jiwen Lu, Peiyang Li, and Jie Zhou.
\newblock Deep progressive reinforcement learning for skeleton-based action
  recognition.
\newblock In {\em CVPR}, pages 5323--5332, 2018.

\bibitem{tang2018mining}
Yansong Tang, Zian Wang, Peiyang Li, Jiwen Lu, Ming Yang, and Jie Zhou.
\newblock Mining semantics-preserving attention for group activity recognition.
\newblock In {\em ACM MM}, pages 1283--1291. ACM, 2018.

\bibitem{DBLP:conf/cvpr/WangNY17}
Minsi Wang, Bingbing Ni, and Xiaokang Yang.
\newblock Recurrent modeling of interaction context for collective activity
  recognition.
\newblock In {\em CVPR}, 2017.

\bibitem{wu2019learning}
Jianchao Wu, Limin Wang, Li Wang, Jie Guo, and Gangshan Wu.
\newblock Learning actor relation graphs for group activity recognition.
\newblock In {\em CVPR}, pages 9964--9974, 2019.

\bibitem{DBLP:journals/corr/abs-1901-00596}
Zonghan Wu, Shirui Pan, Fengwen Chen, Guodong Long, Chengqi Zhang, and
  Philip~S. Yu.
\newblock A comprehensive survey on graph neural networks.
\newblock {\em arXiv preprint arXiv:1901.00596}, 2019.

\bibitem{DBLP:journals/corr/abs-1810-00826}
Keyulu Xu, Weihua Hu, Jure Leskovec, and Stefanie Jegelka.
\newblock How powerful are graph neural networks?
\newblock {\em arXiv preprint arXiv:1810.00826}, 2018.

\bibitem{yan2018participation}
Rui Yan, Jinhui Tang, Xiangbo Shu, Zechao Li, and Qi Tian.
\newblock Participation-contributed temporal dynamic model for group activity
  recognition.
\newblock In {\em ACM MM}, pages 1292--1300. ACM, 2018.

\bibitem{DBLP:conf/aaai/YanXL18}
Sijie Yan, Yuanjun Xiong, and Dahua Lin.
\newblock Spatial temporal graph convolutional networks for skeleton-based
  action recognition.
\newblock In {\em AAAI}, pages 7444--7452, 2018.

\bibitem{DBLP:journals/corr/abs-1810-06543}
Wei Yang, Xiaolong Wang, Ali Farhadi, Abhinav Gupta, and Roozbeh Mottaghi.
\newblock Visual semantic navigation using scene priors.
\newblock {\em arXiv preprint arXiv:1810.06543}, 2018.

\bibitem{DBLP:conf/ijcai/YuYZ18}
Bing Yu, Haoteng Yin, and Zhanxing Zhu.
\newblock Spatio-temporal graph convolutional networks: {A} deep learning
  framework for traffic forecasting.
\newblock In {\em IJCAI}, pages 3634--3640, 2018.

\end{thebibliography}
}

\end{document}